%% file: main.tex
\documentclass[final]{cvpr}

\usepackage{times}
\usepackage{epsfig}
\usepackage{graphicx}
\usepackage{amsmath}
\usepackage{amssymb}
\usepackage{xcolor}
\usepackage{stfloats}
\usepackage{comment}
\usepackage{graphicx}
\usepackage{wrapfig}
\usepackage{caption}

\usepackage[pagebackref=true,breaklinks=true,colorlinks,bookmarks=false]{hyperref}

\def\cvprPaperID{XXXX} 
\def\confYear{CVPR 2021}

\begin{document}

\makeatletter
\newcommand{\thickhline}{%
    \noalign {\ifnum 0=`}\fi \hrule height 1pt
    \futurelet \reserved@a \@xhline
}
\makeatother

\newcommand{\OURS}{SPSG}
\title{\OURS: Self-Supervised Photometric Scene Generation from RGB-D Scans}

\author{
Angela Dai$^1$~~~
Yawar Siddiqui$^1$~~~
Justus Thies$^1$~~~
Julien Valentin$^2$~~~
Matthias Nie{\ss}ner$^1$
\vspace{0.2cm} \\ 
$^1$Technical University of Munich\quad\quad\quad
$^2$Google
\vspace{0.2cm} \\ 
}

\twocolumn[{%
	\renewcommand\twocolumn[1][]{#1}%
	\maketitle
	\begin{center}
		\vspace{-0.2cm}
		\includegraphics[width=0.95\linewidth]{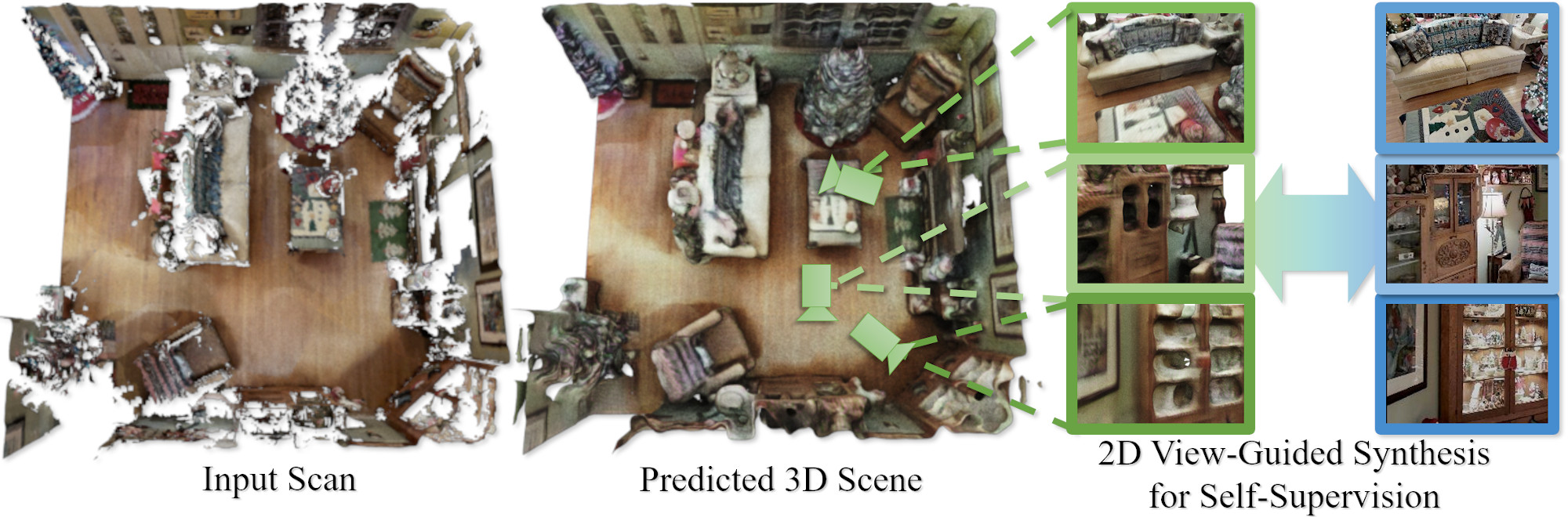}
		\vspace{-0.2cm}
		\captionof{figure}{
		Our \OURS{} approach formulates the problem of generating a complete, colored 3D model from an incomplete scan observation to be self-supervised, enabling training on incomplete real-world scan data. Our key idea is to leverage a 2D view-guided synthesis for self-supervision, comparing rendered views of our predicted model to the original RGB-D frames of the scan.
		Our 2D view-guided synthesis enables outperforming methods relying fully on 3D-based (self-)supervision.
		}
		\label{fig:teaser}
	\end{center}
}]

\maketitle
\thispagestyle{empty}
\pagestyle{empty}

\input{0abstract}

\input{1introduction}

\input{2relatedwork}

\input{3method}

\input{4selfsup_color}

\input{5results}
\input{6conclusion}

\section*{Acknowledgements}
This work was supported by the ZD.B/bidt, a Google Research Grant, a TUM-IAS Rudolf M{\"o}{\ss}bauer Fellowship, an NVidia Professorship Award, the ERC Starting Grant Scan2CAD (804724), and the German Research Foundation (DFG) Grant Making Machine Learning on Static and Dynamic 3D Data Practical.

\newpage
{\small
\bibliographystyle{ieee_fullname}
\bibliography{egbib}
}

\clearpage
\newpage
\begin{appendix}
\input{appendix}

\end{appendix}

\end{document}

%% file: 0abstract.tex
\begin{abstract}
We present \OURS, a novel approach to generate high-quality, colored 3D models of scenes from RGB-D scan observations by learning to infer unobserved scene geometry and color in a self-supervised fashion.
Our self-supervised approach learns to jointly inpaint geometry and color by correlating an incomplete RGB-D scan with a more complete version of that scan.
Notably, rather than relying on 3D reconstruction losses to inform our 3D geometry and color reconstruction, we propose adversarial and perceptual losses operating on 2D renderings in order to achieve high-resolution, high-quality colored reconstructions of scenes.
This exploits the high-resolution, self-consistent signal from individual raw RGB-D frames, in contrast to fused 3D reconstructions of the frames which exhibit inconsistencies from view-dependent effects, such as color balancing or pose inconsistencies.
Thus, by informing our 3D scene generation directly through 2D signal, we produce high-quality colored reconstructions of 3D scenes, outperforming state of the art on both synthetic and real data.
\end{abstract}

%% file: 1introduction.tex
\section{Introduction}

The wide availability of consumer range cameras has propelled research in 3D reconstruction of real-world environments, with applications ranging from content creation to indoor robotic navigation and autonomous driving.
While state-of-the-art 3D reconstruction approaches have now demonstrated robust camera tracking and large-scale reconstruction~\cite{newcombe2011kinectfusion,izadi2011kinectfusion,whelan2015elasticfusion,dai2017bundlefusion}, occlusions and sensor limitation lead these approaches to yield reconstructions that are incomplete both in geometry and in color, making them ill-suited for use in the aforementioned applications.

In recent years, geometric deep learning has made significant progress in learning to reconstruct complete, high-fidelity 3D models of shapes from RGB or RGB-D observations \cite{maturana2015voxnet, dai2017complete, riegler2017OctNet, OccupancyNetworks, Park2019DeepSDFLC}, leveraging synthetic 3D shape data to provide supervision for the geometric completion task.
Recent work has also advanced generative 3D approaches towards operating on larger-scale scenes \cite{song2017ssc,dai2018scancomplete,dai2020sgnn}.
However, producing complete, colored 3D reconstructions of real-world environments remains challenging -- in particular, for real-world observations, we do not have complete ground truth data available. 

We introduce \OURS{}, a generative 3D approach to create high-quality 3D models of real-world scenes from partial RGB-D scan observations in a self-supervised fashion.
Our self-supervised approach leverages incomplete RGB-D scans as target by generating a more incomplete version as input by removing frames. 
This allows correlation of more-incomplete to less-incomplete scans while ignoring unobserved regions.
However, the target scan reconstruction from the given RGB-D scan suffers from inconsistencies in camera alignments and view-dependent effects, resulting in significant color artifacts.
Moreover,  the success of adversarial approaches in 2D image generation~\cite{goodfellow2014generative,karras2017progressive} cannot be directly adopted when the target scan is incomplete, as this results in the `real' examples for the discriminator taking on incomplete characteristics.
Our key observation is that while a 3D scan is incomplete, each individual 2D frame is complete from its viewpoint.
Thus, we leverage the 2D signal provided by the raw RGB-D frames, which provide high-resolution, self-consistent observations as well as photo-realistic examples for adversarial and perceptual losses in 2D.

Thus, our generative 3D model predicts a 3D scene reconstruction represented as a truncated signed distance function with per-voxel colors (TSDF), where we leverage a differentiable renderer to compare the predicted geometry and color to the original RGB-D frames.
In addition, we employ a 2D adversarial and 2D perceptual loss between the rendering and the original input in order to achieve sharp, high-quality, complete colored 3D reconstructions.
Our experiments show that our 2D-based self-supervised approach towards inferring complete geometric and colored 3D reconstructions produces significantly improved performance in comparison to state of state-of-the-art methods, both quantitatively and qualitatively on both synthetic and real data. 

In summary, we present the following contributions:
\begin{itemize}
    \item We introduce the first self-supervised approach to infer a complete, colored reconstruction of 3D scenes from RGB-D scan observations. This enables training solely on incomplete real-world scan data, without requiring domain adaptation from a synthetic regime.
    \item We present a view-based synthesis for differentiable rendering of both TSDF geometry and color, and show that this view-based synthesis outperforms supervision relying on 3D reconstruction of the RGB-D scan data.
\end{itemize}

%% file: 2relatedwork.tex
\section{Related Work}

\paragraph{RGB-D 3D Reconstruction}
3D reconstruction of objects and scenes using RGB-D data is a well explored field \cite{newcombe2011kinectfusion,izadi2011kinectfusion,whelan2015elasticfusion,dai2017bundlefusion}.
For a detailed overview of 3D reconstruction methods, we refer to the state of the art report of \cite{reco_star}.
In addition, our work is related to surface texturing techniques which optimize for texture in observed regions \cite{huang20173dlite,huang2020adversarial}, as well as shading-based refinement approaches which optimize for refined geometry/color in observed regions~\cite{zollhoefer2015shading,maier2017intrinsic3d}. However, in contrast, our goal is to target incomplete scans where color data is missing in the 3D scans.

\paragraph{Learned Single Object Reconstruction}
The reconstruction of single objects given RGB or RGB-D input is an active field of research.
Many works have explored a variety of geometric shape representations, including occupancy grids~\cite{wu20153d}, volumetric truncated signed distance fields~\cite{dai2017complete}, point clouds~\cite{yang2019pointflow}, and recently using deep networks to model implicit surface representations~\cite{Park2019DeepSDFLC, OccupancyNetworks, NIPS2019_8340}.

While such methods have shown impressive geometric reconstruction, generating colored objects has been far less explored.
Im2Avatar~\cite{sun2018im2avatar} predicts an occupancy grid to represent the shape, followed by predicting a color volume.
PIFu~\cite{saito2019pifu} proposes to estimate a pixel-aligned implicit function representing both the shape and appearance of an object, focusing on the reconstruction of humans.
While Texture Fields~\cite{oechsle2019texture} does not reconstruct 3D geometry, this approach predicts the color for a shape by estimating a function mapping a surface position to a color value.
These approaches make significant progress in estimating colored reconstructions, but focus on the limited domain of objects, which are both limited in volume and far more structured than reconstruction of full scenes.

\paragraph{Learned Scene Completion}
While there is a large corpus of work on single object reconstruction, there have been fewer efforts focusing on reconstructing scenes.
SSCNet~\cite{song2017ssc} introduce a method to jointly predict the geometric occupancy and semantic segmentation of a scene from an RGB-D image.
ScanComplete~\cite{dai2018scancomplete} introduces an autoregressive approach to complete partial scans of large-scale scenes.
These approaches focus on geometric and semantic predictions, relying on synthetic 3D data to provide complete ground truth scenes for training, resulting in loss of quality due to the synthetic-real domain gap when applied to real-world scans.
In contrast, SG-NN~\cite{dai2020sgnn} proposes a self-supervised approach for geometric completion of partial scans, allowing training on real data.
Our approach is inspired by that of SG-NN; however, we find that their 3D self-supervision formulation is insufficient for compelling color generation, and instead propose to guide our self-supervision through 2D renderings of our 3D predictions.

%% file: 3method.tex
\section{Method Overview}

Our aim is to generate a complete 3D model, with respect to both geometry and color, from an incomplete RGB-D scan.
We take as input a series of RGB-D frames and estimated camera poses, fused into a truncated signed distance field representation (TSDF) through volumetric fusion~\cite{curless1996volumetric}.
The input TSDF is represented in a volumetric grid, with each voxel storing both distance and color values.
We then learn to generate a TSDF representing the complete geometry and color, from which we extract the final mesh using Marching Cubes~\cite{lorensen1987marching}.

To effectively generate compelling color and geometry for real scan data, we develop a self-supervised approach to learn from incomplete target scans.
From an incomplete target scan, we generate a more incomplete version by removing a subset of its RGB-D frames, and learn the generation process between the two levels of incompleteness while ignoring the unobserved space in the target scan. 
Notably, rather than relying on the incomplete target 3D colored TSDF -- which contains inconsistencies from view-dependent effects, micro-misalignments in camera pose estimation, and is often lower resolution than that of the color sensor (to account for the lower resolution and noise in the depth capture) -- we instead propose a 2D view-guided synthesis, relying on losses formulated on 2D renderings of our predicted TSDF.
As each individual image is self-consistent and high resolution, we mitigate such artifacts by leveraging this image information to guide our predictions.

That is, we render our predicted TSDF to the views of the original images, with which we can then compare our rendered predictions and the original RGB-D frames.
This allows us to exploit the consistency of each individual frame during training, as well as employ not only a reconstruction loss for geometry and color, but also adversarial and perceptual losses, where the `real' target images are the raw RGB-D frames.
Each of these views is complete, high-resolution, and photo-realistic, which provides guidance for our approach to learn to generate complete, high-quality, colored 3D models.

%% file: 4selfsup_color.tex
\section{Self-supervised Photometric Generation}

The key idea of our method for photometric scene generation from incomplete RGB-D scan observations is to formulate  a self-supervised approach based on 2D view-guided synthesis, leveraging rendered views of our predicted 3D model.
Since training on real-world scan data is crucial for realistic color generation, we need to be able to learn from incomplete target scan data as complete ground truth is unavailable for real-world scans.

Thus, we learn a generative process from the correlation of an incomplete target scan composed of RGB-D frames $\{f_k\}$ with a more incomplete version of that scan constructed from a subset of the frames $\{f_i\}\subset \{f_k\}$.
The input scan $\mathcal{S}_i$ during training is then created by volumetric fusion of $\{f_i\}$ to a volumetric TSDF with per-voxel distances and colors.
This is inspired by the SG-NN approach~\cite{dai2020sgnn}; however, crucially, rather than relying on the fused incomplete target TSDF, we formulate 2D-based rendering losses to guide both geometry as well as color prediction.
This both avoids smaller-scale artifacts from inconsistencies in camera pose estimation as well as view-dependent lighting and color balancing, and importantly, allows formulation of adversarial and perceptual losses with the raw RGB-D frames, which are individually complete views in image space.
These losses are critical towards producing compelling photometric scene generation results.

Additionally, our self-supervision exploits the different patterns of incompleteness seen across a variety of target scans, where each individual target scan remains incomplete but learning across a diverse set of patterns enables generating output 3D models that have more complete, consistent geometry and color than any single target scan seen during training.

\subsection{Differentiable Rendering}

To formulate our 2D-based losses, we render our predicted TSDF $\mathcal{S}_p$ in a differentiable fashion, generating color, depth, and world-space normal images, $C_v$, $D_v$, and $N_v$, for a given view $v$.
We then operate on $C_v$, $D_v$ and $N_v$ to formulate our reconstruction, adversarial, and perceptual losses.

\begin{figure}
	\centering
	\includegraphics[width=0.9\linewidth]{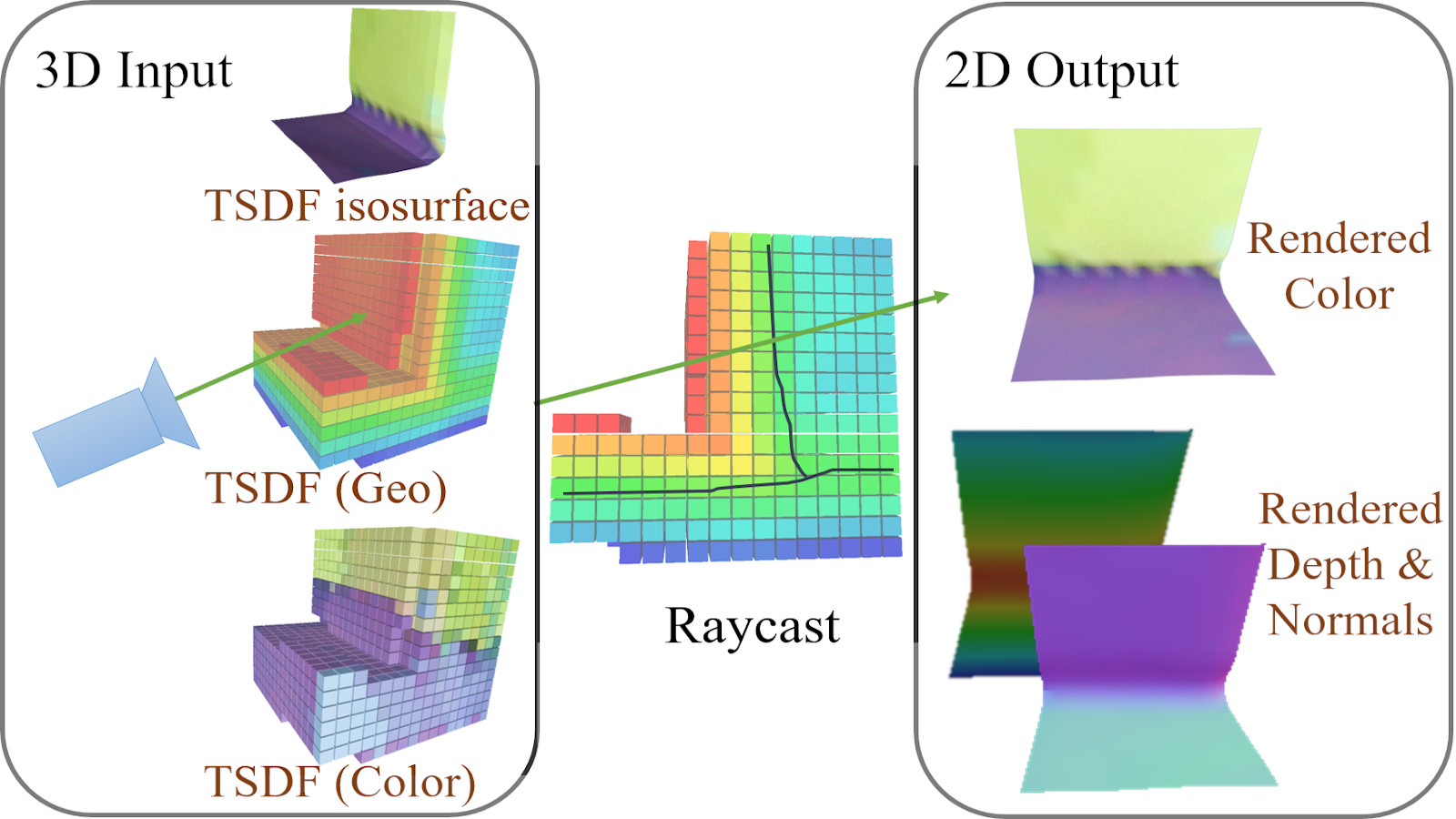}
	\vspace{-0.3cm}
	\caption{
	    Differentiable rendering of our 3D predicted TSDF geometry and color.
    }
	\label{fig:losses2d}
\end{figure}

Specifically, for $\mathcal{S}_p$ comprising per-voxel distances and colors, and a camera view $v$ with the intrinsics (focal length, principal point), extrinsics (rotation, translation), and image dimensions, we generate $C_v$, $D_v$, and $N_v$ by raycasting, as shown in Figure~\ref{fig:losses2d}.
For each pixel in 
 the output image, we construct a ray $r$ from the view $v$ and march along $r$ through $\mathcal{S}_p$ using trilinear interpolation to determine TSDF values.
To locate the surface at the zero-crossing of $\mathcal{S}_p$, we look for sign changes between current and previous TSDF values.

For efficient search, we first use a fixed increment to search along the ray (half of the truncation value), and once a zero-crossing has been detected, we use an iterative line search to refine the estimate. 
The refined zero-crossing location is then used to provide the depth, normal, and color values for $D_v$, $N_v$, and $C_v$ as distance from the camera, negative gradient of the TSDF, and associated color value, respectively.

Our differentiable TSDF rendering is implemented in CUDA as a PyTorch extension for efficient runtime, with the backward pass similarly implemented through ray marching, using atomic add operations to accumulate gradient information when multiple pixels correspond to a voxel.

\subsection{2D View-Guided Synthesis / Re-rendering loss}

Our self-supervised approach is based on 2D losses operating on the depth, normal, and color images $D_v$, $N_v$, and $C_v$, which are rendered from the predicted TSDF $\mathcal{S}_p$.
This enables comparison to the original RGB-D frame data $D^{t}_v$, $N^{t}_v$ (normals are computed in world space from the depth images), and $C^{t}_v$, thus, avoiding explicit view inconsistencies in the targets as well as providing complete target view information.
For the task of generating a complete photometric reconstruction from an incomplete scan, we employ a reconstruction loss to anchor geometry and color predictions, as well as an adversarial and perceptual loss, to capture more realistic appearance in the final prediction.

\paragraph{Reconstruction Loss.}

We use an $\ell_1$ loss to guide depth and color to the target depth and color:
\begin{align*}
    L^R_D &= \frac{1}{N}\sum_p||D_v(p) - D^t_v(p)||_1 \\
    L^R_C &= \frac{1}{3N}\sum_p||C_v(p) - C^t_v(p)||_1.
\end{align*}
Since the rendered $D_v$ and $C_v$ may not have valid values for all pixels (where no surface geometry was seen), these losses operate only on the valid pixels $p$, normalized by  the number of valid pixels $N$. 
The color loss operates on the 3 channels of the CIELAB color space, which we empirically found to provide better color performance than RGB space.
Note that these reconstruction losses as formulated have a trivial solution where generating no surface geometry in $\mathcal{S}_p$ provides no loss, so we employ an $\ell_1$ 3D geometric reconstruction loss $L^R_G$ on the predicted 3D TSDF distances, weighted by a small value $w_g$ to discourage lack of surface geometry prediction.
For $L^R_G$, we mask out any voxels which were unobserved in the target scan.
The final reconstruction loss is then $L^R = w_gL^R_G + L^R_D + L^R_C$.

\paragraph{Adversarial Loss.}

To capture a more realistic photometric scene generation, we employ an adversarial loss on both $N_v$ and $C_v$. Note that since depth values can vary dramatically for the same geometry from different views, we consider the world-space normals $N_v$ rather than $D_v$ for the adversarial loss. 
In particular, this helps avoid averaging artifacts when only the reconstruction loss is used, which helps markedly in addressing color imbalance in the training set (e.g., color dominated by walls/floors colors which typically have little diversity).
We use the conditional adversarial loss:
\begin{align*}
    L^A = &\mathbb{E}_{x,N_v,C_v}(\log D(x,[N_v,C_v])) + \\
    &\mathbb{E}_{x,N^t_v,C^t_v}(\log (1 - D(x, [N^t_v,C^t_v] ) )
\end{align*}
where $[\cdot,\cdot]$ denotes concatenation, and $x$ is the condition, with $x = [N^i_v,C^i_v]$ where $N^i_v,C^i_v$ are the rendered normal and color images of the input scan $\mathcal{S}_i$ from view $v$.
Note that although $N^t_v$ and $C^t_v$ can be considered complete in the image view,  $N_v$ and $C_v$ may contain invalid pixels; for these invalid pixels we copy the corresponding values from $D^t_v$ and $C^t_v$ to avoid trivially recognizing real from synthesized by number of invalid pixels.

Similar to Pix2Pix~\cite{isola2017image}, we use a patch-based discriminator, on $94\times94$ patches of $320\times 256$ images.

\paragraph{Perceptual Loss.}

We additionally employ a loss to penalize perceptual differences from the rendered color images of our predicted TSDF.
We use a pretrained VGG network~\cite{simonyan2014very}, and use a content loss~\cite{gatys2016image} where feature maps from the eighth convolutional layer are compared with an $\ell_2$ loss.
\begin{equation*}
    L^P = ||\textrm{VGG}_8(C_v) - \textrm{VGG}_8(C^t_v)||_2
\end{equation*}

{
\begin{table*}[bp]
\begin{center}
	\small
	\begin{tabular}{| l || c | c | c | c |}
		\hline
		Method &  SSIM ($\uparrow$) & Feature-$\ell_1$ ($\downarrow$) & FID ($\downarrow$) \\  \thickhline
        PIFu$^+$~\cite{saito2019pifu} & 0.67 & 0.25 & 81.5  \\ \hline
        Texture Fields~\cite{oechsle2019texture} (on Ours Geometry) & 0.70 & 0.23 & 68.4  \\ \hline
        Ours & {\bf 0.71} & {\bf 0.22} & {\bf 56.0}  \\ \hline
	\end{tabular}
	\vspace{-0.3cm}
	\caption{Evaluation of colored reconstruction from incomplete scans of Matterport3D~\cite{Matterport3D} scenes. We evaluate rendered views of the outputs of all methods against the original color images. 
	}
	\label{tab:color_matterport}		
\end{center}
\end{table*}
}

\subsection{Data Generation}

To generate the input and target scans $\mathcal{S}_i$ and $\mathcal{S}_t$ used during training, we use a random subset of the target RGB-D frames (in our experiments, $50\%$ ) to construct $\mathcal{S}_i$.
Both $\mathcal{S}_i$ and $\mathcal{S}_t$ are then constructed through volumetric fusion~\cite{curless1996volumetric}; we use a voxel resolution of $2$cm.
In order to realize efficient training, we train on cropped chunks of the input-target pairs of size $64\times 64\times 128$ voxels.
These chunks are sampled uniformly, while discarding chunks with less than $0.5\%$ geometric occupancy.
For each train chunk, we associate up to five RGB-D frames based on their geometric overlap with the chunk, using frames with the most IoU with the back-projected frame. 
These frames are used as targets for the 2D losses on the rendered predictions.

\subsection{Network Architecture}

\begin{figure*}
	\centering
	\includegraphics[width=0.95\linewidth]{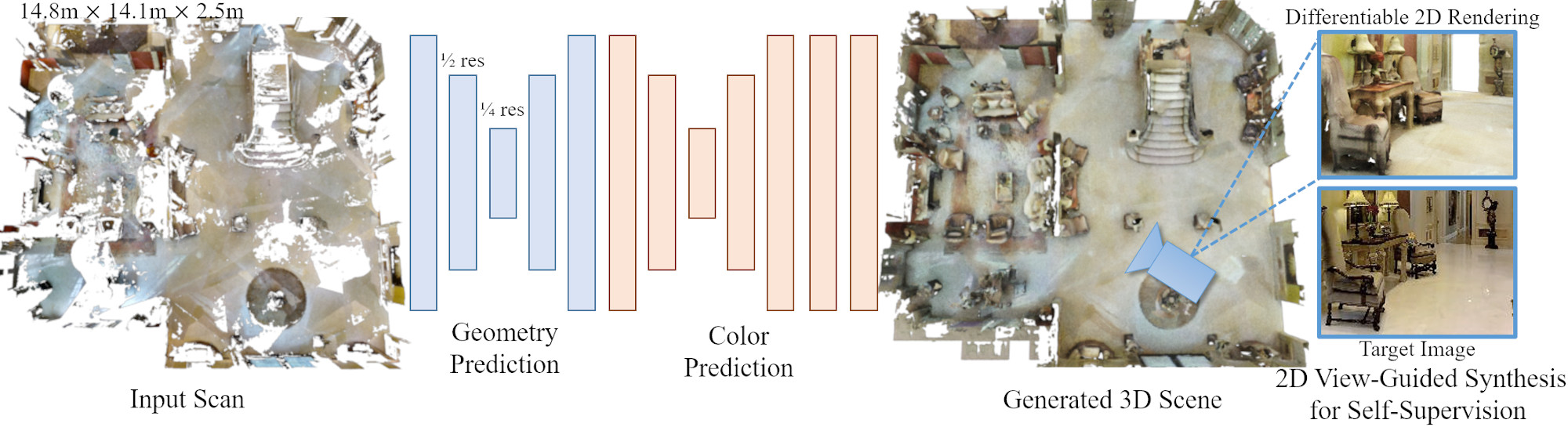}
	\vspace{-0.3cm}
	\caption{
	    Network architecture overview. Our approach is fully-convolutional, operating on an input TSDF volume and predicting an output TSDF, from which we apply our 2D view-guided synthesis.
    }
	\label{fig:architecture}
\end{figure*}

Our network, visualized in Figure~\ref{fig:architecture}, is designed to produce a 3D volumetric TSDF representation of a scene from an input volumetric TSDF.
We predict both geometry and color in a fully-convolutional, end-to-end-trainable fashion.
We first predict geometry, followed by color, so that the color predictions can be directly informed by the geometric structure.
The geometry is predicted with an encoder-decoder structure, then color using an encoder-decoder followed by a series of convolutions which maintain spatial resolution. 

The encoder-decoder for geometry prediction spatially subsamples to a factor $1/4$ of the original resolution, and outputs a feature map $f_g$ from which the final geometry is predicted.
The geometric predictions then inform the color prediction, with $f_g$ input to the next encoder-decoder.
The color prediction is structured similarly to the geometry encoder-decoder, with a series of additional convolutions maintaining the spatial resolution.
We found that avoiding spatial subsampling before the color prediction helped to avoid checkering artifacts in the predicted color outputs.

Our discriminator architecture is composed of a series of 2D convolutions, each spatially subsampling its input by a factor of 2.
For a detailed architecture specification, we refer to the appendix.

\paragraph{Training Details}

We train our approach on a single NVIDIA GeForce RTX 2080.
We weight the loss term $L^R_G$ with $w_g=0.1$ and the adversarial loss for the generator by $0.005$; all other terms in the loss have a weight of $1.0$.
We use the Adam optimizer with a learning rate of $0.0001$ and batch size of $2$, and train our model for $\approx 48$ hours until convergence.
For efficient training, we train on $64\times 64\times 128$ cropped chunks of scans; at test time, since our model is fully-convolutional, we operate on entire incomplete scans of varying sizes as input.

%% file: 5results.tex
\section{Results}

{
\begin{table*}[tp]
\begin{center}
	\small
	\begin{tabular}{| l || c | c | c | c |}
		\hline
		Method &  SSIM ($\uparrow$) & Feature-$\ell_1$ ($\downarrow$) & FID ($\downarrow$) \\  \thickhline
        Baseline-3D & 0.694 & 0.236 & 80.51  \\ \hline 
        Ours ($\ell_1$ only) & 0.699 & 0.231 & 67.92  \\ \hline 
        Ours (no adversarial) & 0.695 & 0.229 & 62.15  \\ \hline 
        Ours (no perceptual) & 0.699 & 0.227 & 61.46  \\ \hline \hline
        Ours & {\bf 0.709} & {\bf 0.219} & {\bf 56.03}  \\ \hline
	\end{tabular}
	\vspace{-0.3cm}
	\caption{Ablation study of our design choices on Matterport3D~\cite{Matterport3D} scans. 
	}
	\label{tab:ablation}		
\end{center}
\end{table*}
}

To evaluate our \OURS{} approach, we consider the real-world scans from the Matterport3D dataset~\cite{Matterport3D}, where no complete ground truth is available for color and geometry, and additionally provide further analysis on synthetic data from the chair class of ShapeNet~\cite{shapenet2015}, where complete ground truth data is available.
To enable quantitative evaluation on Matterport3D scenes, we consider input scans generated with $50\%$ of all available RGB-D frames for each scene, and evaluate against the target scan composed of all available RGB-D frames (ignoring unobserved space).
For ShapeNet, we consider single RGB-D frame input, and the complete shape as the target.

\paragraph{Evaluation metrics}
To evaluate our color reconstruction quality, we adopt several metrics to evaluate rendered views of the predicted meshes in comparison to the original views (as we do not have complete 3D color data available for real-world scenarios).
The structure similarity image metric (SSIM)~\cite{brunet2011mathematical} is often used to measure more local characteristics in comparing a synthesized image directly to the target image, but can tend to favor averaging over sharp detail.
We also use a perceptual metric, Feature-$\ell_1$, following the metric proposed in \cite{oechsle2019texture}, which evaluates the $\ell_1$ distance between the feature embeddings of the synthesized and target images under an InceptionV3 network~\cite{szegedy2016rethinking}.
Finally, we consider the Fr\'echet Inception Distance (FID)~\cite{heusel2017gans}, which is commonly used to evaluate the quality of images synthesized by 2D generative techniques, and captures a distance between the distributions of synthesized images and real images.

To measure the geometric quality of our reconstructed shapes and scenes, we use  intersection-over-union (IoU) and Chamfer distance. 
IoU is computed over the voxelization of the output meshes of all approaches, with voxel size of $2$cm for Matterport3D data and $0.01$ (relative to the unit normalized space) for ShapeNet data.
For Chamfer distance, we sample 30K points from the output meshes as well as ground truth meshes, and compute the distance in metric space for Matterport3D and normalized space for ShapeNet.
Note that for the case of real scans, we estimate unobserved space in the target and ignore it for the geometric evaluation; additionally, since the unobserved space is estimated, we also evaluate recall as the amount of intersection of the prediction with the target divided by the target.

For all comparisons to state-of-the-art approaches predicting both color and geometry, we provide as input the incomplete TSDF and color, and if necessary, adapt the method's input (denoted by $^+$). All other training schemes for state-of-the-art comparisons follow that of the proposed approach, trained on our generated data from Matterport3D scenes and ShapeNet objects.

{
\begin{table*}[tp]
\begin{center}
	\small
	\vspace{-0.3cm}
	\begin{tabular}{| l || c | c | c | c |}
		\hline
		Method &  SSIM ($\uparrow$) & Feature-$\ell_1$ ($\downarrow$) & FID ($\downarrow$) \\  \thickhline
		Im2Avatar~\cite{sun2018im2avatar} & 0.85 & 0.25 & 59.7  \\ \hline
        PIFu$^+$~\cite{saito2019pifu} & 0.86 & 0.24 & 70.3  \\ \hline
        Texture Fields~\cite{oechsle2019texture} (on Ours Geometry) & {\bf 0.93} & 0.20 & 30.3  \\ \hline
        Ours & {\bf 0.93} & {\bf 0.19} & {\bf 29.0}  \\ \hline
	\end{tabular}
	\vspace{-0.3cm}
	\caption{Evaluation of colored reconstruction from incomplete scans of ShapeNet~\cite{shapenet2015} chairs.
	}
	\label{tab:color_shapenet}		
\end{center}
\end{table*}
}
\begin{figure*}[tp]
	\centering
	\includegraphics[width=0.95\linewidth]{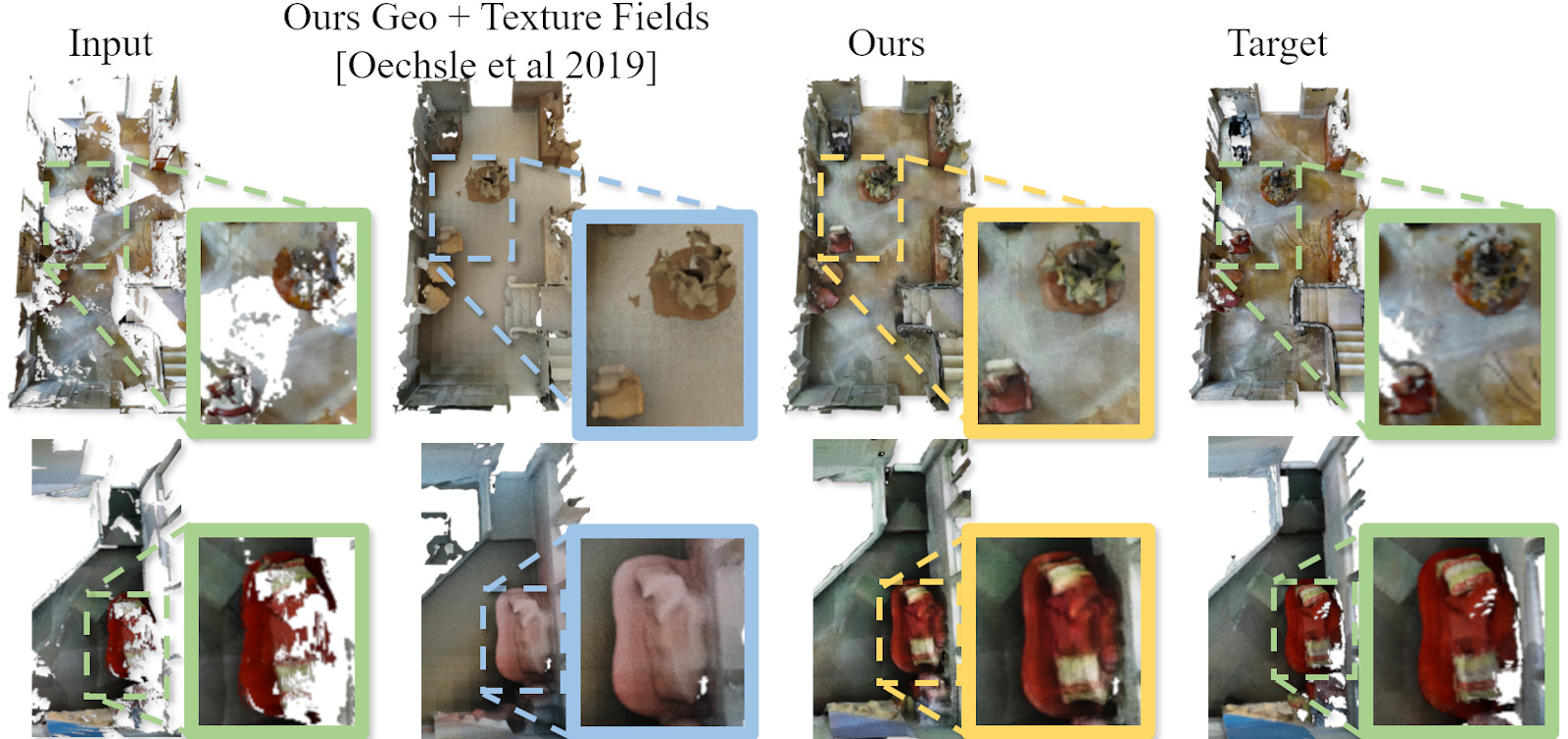}
	\vspace{-0.1cm}
	\caption{
	    Qualitative evaluation of colored reconstruction on Matterport3D~\cite{Matterport3D} scans.
    }
	\label{fig:cmp_matterport}
\end{figure*}

\paragraph{Self-supervised photometric scene generation.}
We demonstrate our self-supervised approach to generate reconstructions of scenes from incomplete scan data, using scan data from Matterport3D~\cite{Matterport3D} with the official train/test split (72/18 trainval/test scenes comprising 1788/394 rooms).
Tables~\ref{tab:color_matterport}  and \ref{tab:geo} show a comparison of our approach to state-of-the-art methods for color and geometry reconstruction: PIFu~\cite{saito2019pifu}, Texture Fields~\cite{oechsle2019texture}, and SG-NN~\cite{dai2020sgnn}.
Here, our view-guided losses enable more effective generation that the 3D-only target supervision used for PIFu, Texture Fields, and SG-NN, as it helps to mitigate learning from artifacts (e.g., small-scale camera misalignment) embedded in the reconstructed 3D target.
Note that since Texture Fields predicts only color, we provide our predicted geometry as input; for test scenes, since it is designed for fixed volume sizes, we apply it in sliding window fashion.
We additionally show qualitative results in Figure~\ref{fig:cmp_matterport}, as well as qualitative geometric comparisons in the appendix.
All methods were trained on the generated input-target pairs of scans from Matterport3D with frames removed from the target scan to create the corresponding inputs, and the respective proposed loss functions used for training.

For color reconstruction, PIFu~\cite{saito2019pifu} and Texture Fields~\cite{oechsle2019texture} capture the coarse structure of the scene colors, but the complexity of local detail in the scenes is often lost, while our 2D-guided losses incorporating perceptual components enable capturing a more realistic color distribution.
For geometric reconstruction, our emphasis on 2D guidance in contrast to the 3D guidance of PIFu~\cite{saito2019pifu}, OccNet~\cite{OccupancyNetworks}, and SG-NN~\cite{dai2020sgnn} mitigates learning from artifacts in the fused 3D reconstruction of the scenes (e.g., from small camera estimation errors), producing a more effective geometric scene generation.

We additionally show that training only on synthetic data is insufficient for color and geometry generation for real-world scene data; as SG-NN~\cite{dai2020sgnn} trained on the synthetic scene dataset of \cite{song2017ssc} results in worse performance than training on real-world Matterport3D data, due to the domain gap between synthetic and real scenes.

\begin{figure}
	\centering
	\includegraphics[width=\linewidth]{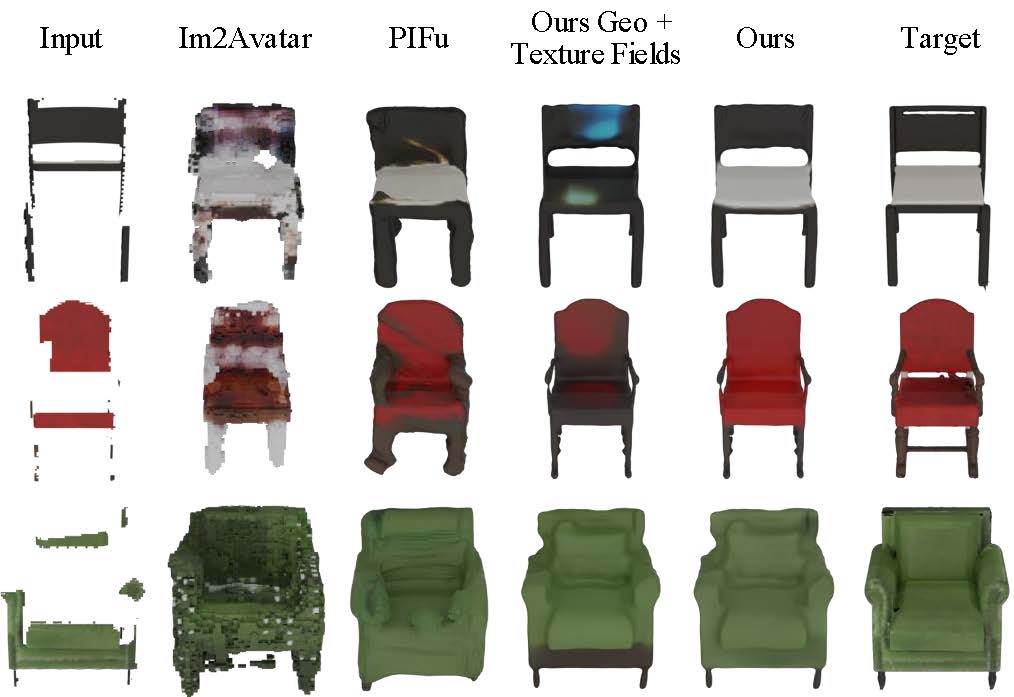}
	\vspace{-0.5cm}
	\caption{
	Colored reconstruction on ShapeNet~\cite{shapenet2015} chairs, in comparison with Im2Avatar~\cite{sun2018im2avatar}, PIFu~\cite{saito2019pifu}, and Texture Fields~\cite{oechsle2019texture} (run on geometry predicted by our method).
    }
	\label{fig:cmp_shapenet}
\end{figure}

\begin{figure*}
	\centering
	\includegraphics[width=0.9\linewidth]{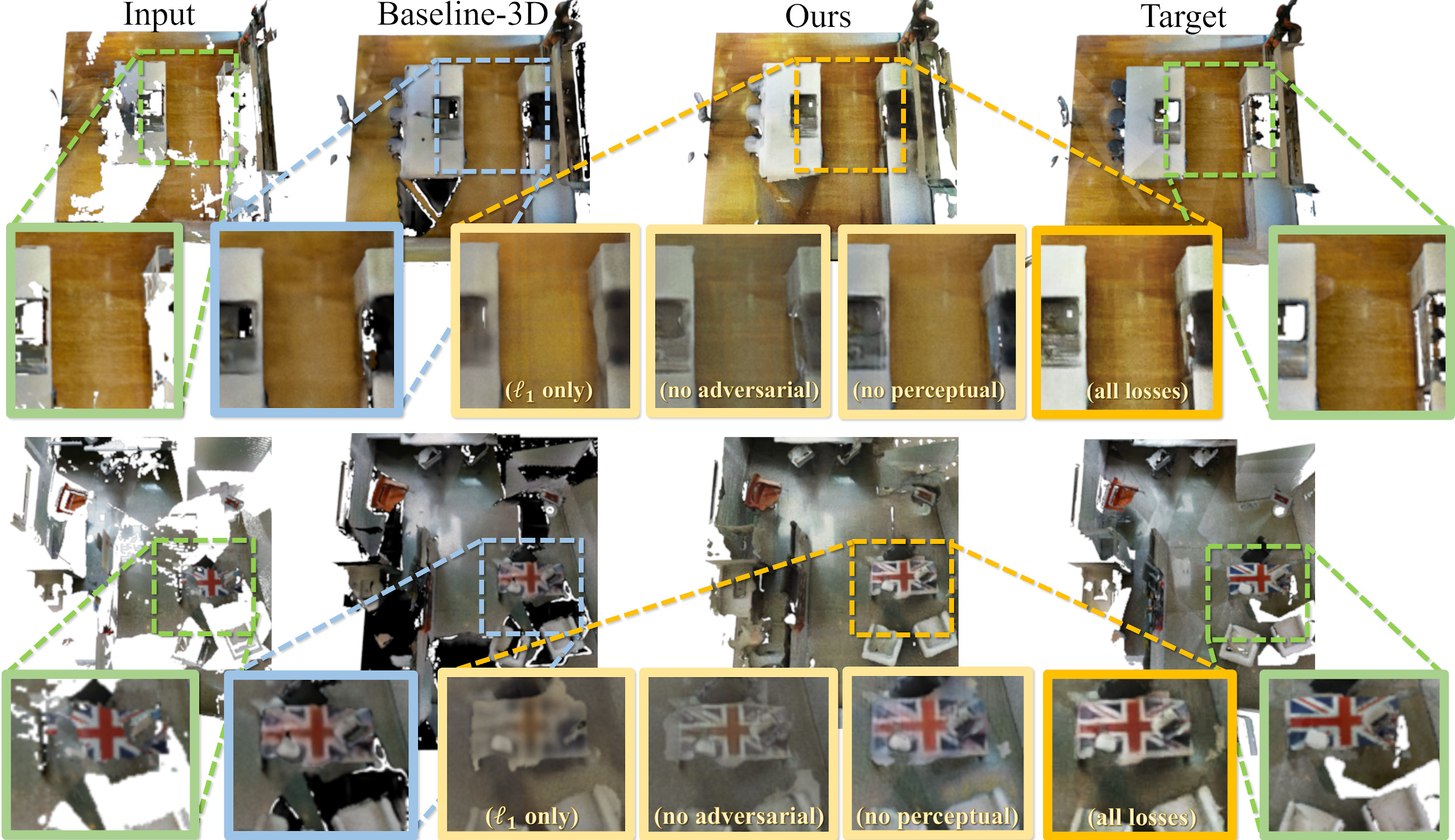}
	\vspace{-0.2cm}
	\caption{
	Qualitative evaluation of our design choices on Matterport3D~\cite{Matterport3D} scans.
    }
	\label{fig:ablation}
\end{figure*}

\paragraph{What is the effect of the 2D view-guided synthesis?}
In Table~\ref{tab:ablation}, we analyze the effects of our various 2D rendering based losses, and show qualitative results in Figure~\ref{fig:ablation}.
We first replace our rendering-based losses with analogous 3D losses, i.e., $L^R, L^A$, and $L^P$ use the 3D incomplete target TSDF instead of 2D views (\emph{Baseline-3D}). This approach learns to reflect the inconsistencies present in the fused 3D target scan (e.g., striping artifacts where one frame ends and another begins), and moreover, suffers from the incompleteness of the target scan data when used as `real' examples for the discriminator and the perceptual loss (causing black artifacts in some missing regions).
Thus, our approach to leverage rendering-based losses with original RGB-D frames produces more consistent, compelling reconstructions.

Additionally, we evaluate the effect of our adversarial and perceptual losses on the output color quality, evaluating our approach with the adversarial loss removed (\emph{Ours (no adversarial)}), perceptual loss removed (\emph{Ours (no perceptual)}), and both adversarial and perceptual losses removed (\emph{Ours ($\ell_1$ only)}).
Using only an $\ell_1$ loss results in blurry, washed out colors.
With the adversarial loss, the colors are less washed out, and with the perceptual loss, colors become sharper; using all losses combines these advantages to achieve compelling scene generation.

\begin{figure}
	\centering
	\includegraphics[width=0.8\linewidth]{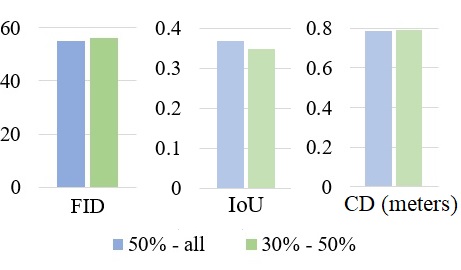}
	\vspace{-0.3cm}
	\caption{
	Effect of varying incompleteness of target data during training on geometry and color for Matterport3D~\cite{Matterport3D}.
    }
	\label{fig:varying_incompleteness}
\end{figure}

{
\begin{table*}
\begin{center}
	\small
	\begin{tabular}{| l || c | c | c |  }
	    \multicolumn{4}{c}{Matterport3D} \\
		\hline
		Method &  IoU ($\uparrow$) & Recall ($\uparrow$) & Chamfer Dist. ($\downarrow$) \\  \thickhline
        OccNet$^+$~\cite{OccupancyNetworks} & 0.05 & 0.13 & 0.16 \\ \hline
        PIFu$^+$~\cite{saito2019pifu} & 0.06 & 0.34 & 0.05  \\ \hline
        ConvOccNet~\cite{peng2020convolutional} & 0.24 & 0.48 & 0.03 \\ \hline
        SG-NN~\cite{dai2020sgnn} (synth) & 0.27 & 0.55 & 0.03  \\ \hline
        SG-NN~\cite{dai2020sgnn} & 0.28 & 0.57 & 0.02  \\ \hline
        Baseline-3D & 0.33 & 0.58 & 0.04  \\ \hline
        Ours & {\bf 0.39} & {\bf 0.64}  & {\bf 0.01}   \\ \hline
	\end{tabular}
	\quad
	\begin{tabular}{| l || c | c | }
	    \multicolumn{3}{c}{ShapeNet} \\
		\hline
		Method &  IoU ($\uparrow$) & Chamfer Dist. ($\downarrow$) \\  \thickhline
		Im2Avatar~\cite{sun2018im2avatar} & 0.17 & 0.27 \\ \hline
        PIFu$^+$~\cite{saito2019pifu} & 0.34 & 0.27  \\ \hline
        OccNet$^+$~\cite{OccupancyNetworks} & 0.46 & 0.20  \\ \hline
        Ours & {\bf 0.66} & {\bf 0.09}   \\ \hline
	\end{tabular}
	\vspace{-0.2cm}
	\caption{Evaluation of geometric reconstruction from  Matterport3D~\cite{Matterport3D} scans (left) and  ShapeNet~\cite{shapenet2015} chairs (right).
	}
	\label{tab:geo}
\end{center}
\end{table*}
}

\paragraph{Effect of varying levels of corruption.}
In Figure~\ref{fig:varying_incompleteness}, we show the effect of varying degrees of completeness of the target data used during training on both color and geometry reconstruction. We compare using all target data available (denoted in blue) with 50\% of the target frames (denoted in green); our performance remains robust even under this degradation.

\paragraph{Evaluation on synthetic 3D shapes.}
We additionally evaluate our approach in comparison to state-of-the-art methods on synthetic 3D data, using the chairs category of ShapeNet (5563/619 trainval/test shapes).
All methods are provided a single RGB-D frame as input, and for training, the complete shape as target. 
Tables \ref{tab:geo} and \ref{tab:color_shapenet} show quantitative evaluation for geometry and color predictions, respectively. 
Our approach predicts more accurate geometry, and our adversarial and perceptual losses provide more compelling color generation.

\paragraph{Limitations.}
Our approach shows promising results for simultaneous color and geometry scene generation from real-world observations; however, for a more realistic appearance model, incorporating estimation of lighting and material properties beyond a diffuse assumption is required.
Additionally, our volumetric 3D approach can also be limited in extending to very high resolutions (e.g., sub-millimeter) for very fine-grained color modeling.

%% file: 6conclusion.tex
\section{Conclusion}

We introduce \OURS{}, a self-supervised approach to generate complete, colored 3D models from incomplete RGB-D scan data.
Our 2D view-guided formulation enables self-supervision as well as compelling color generation through 2D adversarial and perceptual losses.
Thus we can train and test on real-world scan data where complete ground truth is unavailable, avoiding the large domain gap in using synthetic color and geometry data.
We believe this is an exciting avenue for future research, and provides an interesting alternative for synthetic data generation or domain transfer.

%% file: appendix.tex
\section{Network Architecture}

\begin{figure*}[h]
	\centering
	\includegraphics[width=0.7\linewidth]{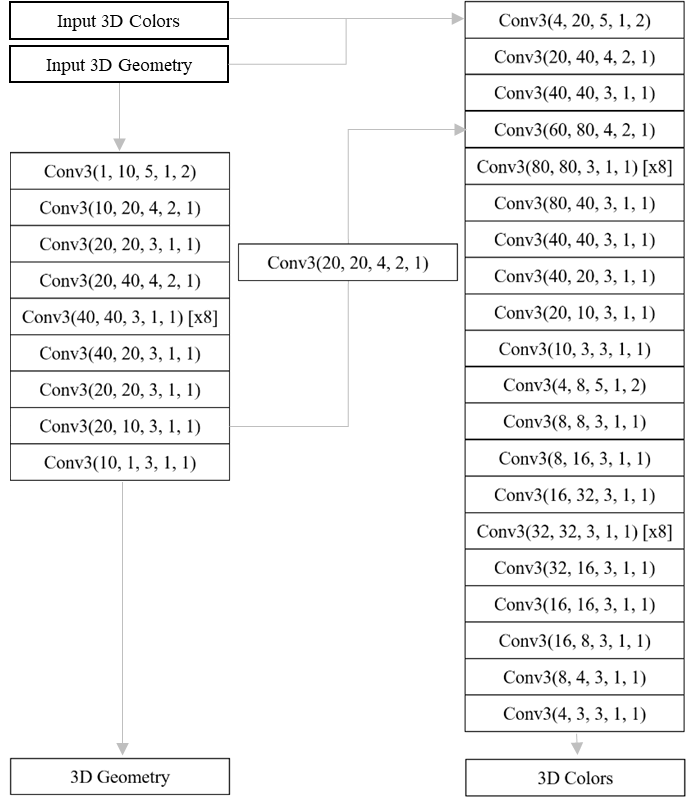}
	\caption{
	Network architecture specification. Given an incomplete RGB-D scan, we take its 3D geometry and color as input, and leverage a fully-convolutional neural network to predict the complete 3D model represented volumetrically for both geometry and color.
    }
	\label{fig:architecture_spectification}
\end{figure*}

We detail our network architecture specifications in Figure~\ref{fig:architecture_spectification}.
Convolution parameters are given as (nf\_in, nf\_out, kernel\_size, stride, padding).
Each convolution (except those producing final outputs for geometry and color) is followed by a Leaky ReLU and batch normalization.

\begin{figure*}[tp]
	\centering
	\includegraphics[width=\linewidth]{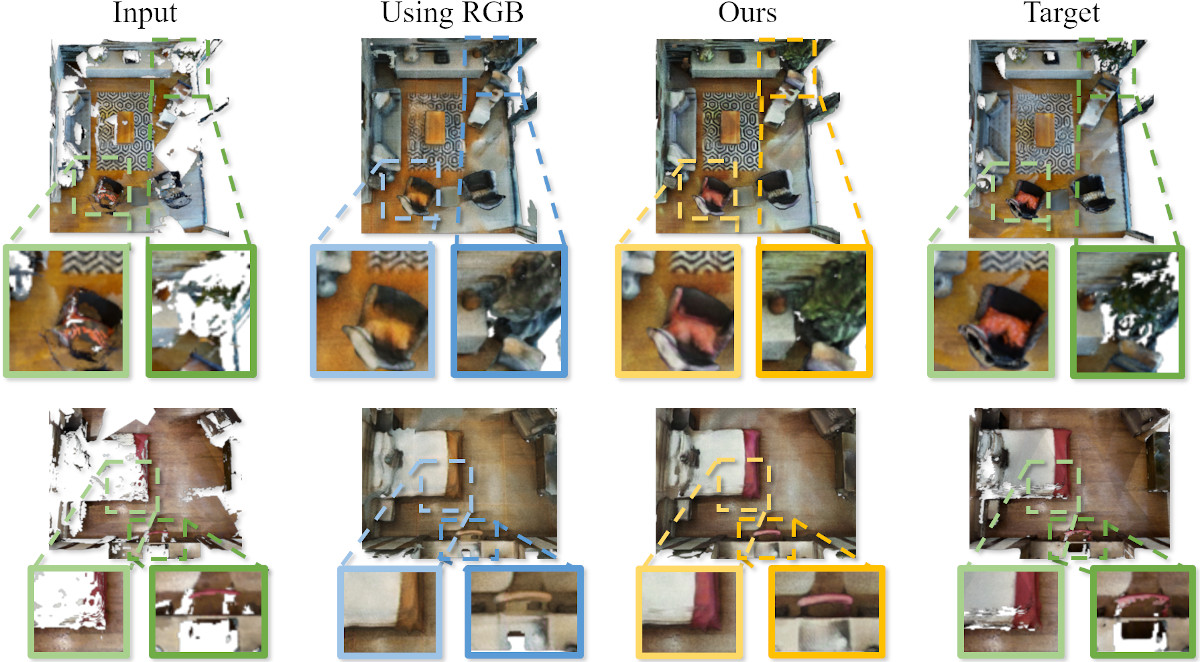}
	\caption{
	Qualitative comparison of our approach using CIELAB color space vs RGB color space on  Matterport3D~\cite{Matterport3D} scans.
	Using CIELAB space allows us to capture more diversity in output color generation.
    }
	\label{fig:lab_vs_rgb}
\end{figure*}

\section{Additional Results}

\subsection{Additional Ablation Studies}

We additionally evaluate the effect of the CIELAB color space that our approach uses for color generation, in comparison to RGB space. 
Table~\ref{tab:lab_vs_rgb} quantitatively evaluates the color generation, showing that CIELAB space is more effective, and Figure~\ref{fig:lab_vs_rgb} shows that using CIELAB space allows our approach to capture a greater diversity of colors in our output predictions.

We also evaluate the geometric reconstruction when trained with a 3D $\ell_1$ loss only in Table~\ref{tab:geo_ab}; here, Baseline-3D can improve on an $\ell_1$ loss only, and ours leverages the advantages of a view-guided synthesis for the best reconstruction performance.

{
\begin{table*}[h]
\begin{center}
	\small
	\begin{tabular}{| l || c | c | c | c |}
		\hline
		Method &  SSIM ($\uparrow$) & Feature-$\ell_1$ ($\downarrow$) & FID ($\downarrow$) \\  \thickhline
        Using RGB & 0.702 & 0.222 & 58.8  \\ \hline 
        Ours & {\bf 0.709} & {\bf 0.219} & {\bf 56.03}  \\ \hline
	\end{tabular}
	\caption{Comparison of our approach using CIELAB color space to using RGB on Matterport3D~\cite{Matterport3D} scans. CIELAB produces more effective color generation.
	}
	\label{tab:lab_vs_rgb}		
\end{center}
\end{table*}
}4

{
\begin{table*}
\begin{center}
	\small
	\begin{tabular}{| l || c | c | c |  }
	    \multicolumn{4}{c}{Matterport3D} \\
		\hline
		Method &  IoU ($\uparrow$) & Recall ($\uparrow$) & Chamfer Dist. ($\downarrow$) \\  \thickhline
        3D $\ell_1$ loss only & 0.31 & 0.58 & 0.02  \\ \hline
        Baseline-3D & 0.33 & 0.58 & 0.04  \\ \hline
        Ours & {\bf 0.39} & {\bf 0.64}  & {\bf 0.01}   \\ \hline
	\end{tabular}
	\vspace{-0.2cm}
	\caption{Additional ablations on geometric reconstruction from  Matterport3D~\cite{Matterport3D} scans.
	}
	\label{tab:geo_ab}
\end{center}
\end{table*}
}

\begin{figure*}[bp]
	\centering
	\includegraphics[width=\linewidth]{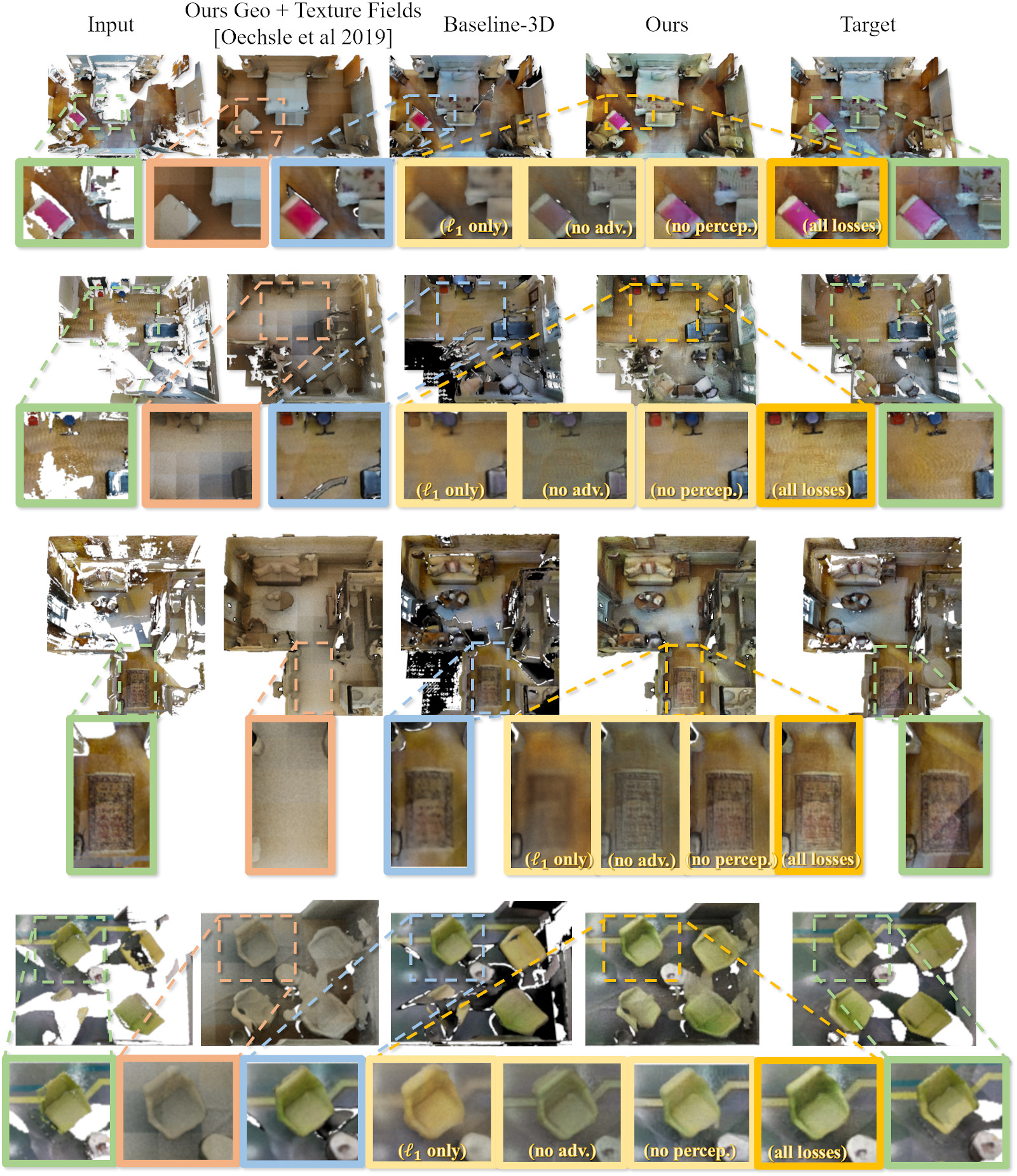}
	\caption{
	Additional qualitative evaluation of colored reconstruction on Matterport3D~\cite{Matterport3D} scans.
    }
	\label{fig:cmp_mp_more}
\end{figure*}

\begin{figure*}
	\centering
	\includegraphics[width=0.75\linewidth]{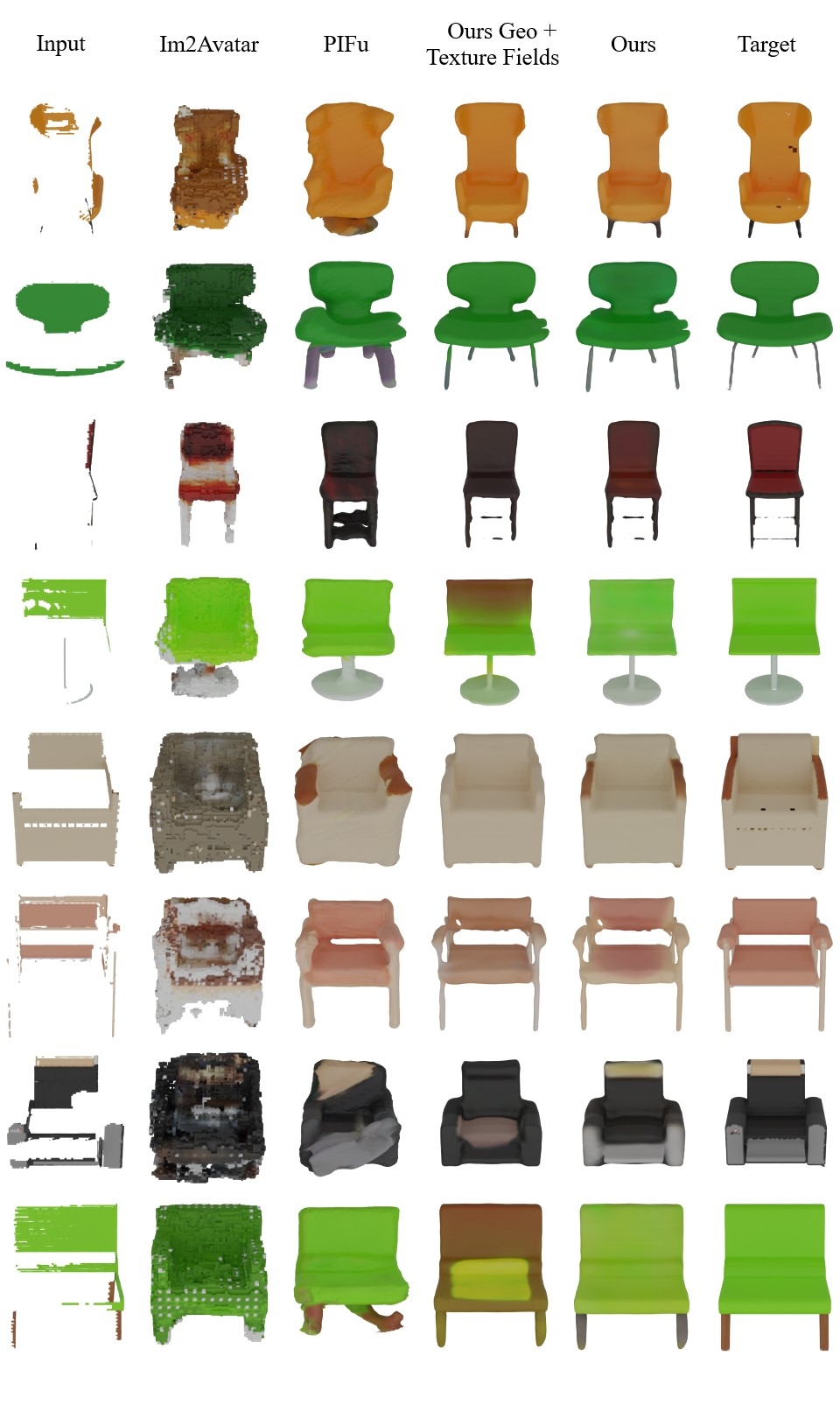}
	\vspace{-0.3cm}
	\caption{
	    Additional qualitative evaluation of colored reconstruction of our method against Im2Avatar~\cite{sun2018im2avatar}, PIFu~\cite{saito2019pifu}, and Texture Fields~\cite{oechsle2019texture} (run on geometry predicted by our method) on ShapeNet~\cite{shapenet2015} chairs.
    }
	\label{fig:cmp_shapenet_more}
\end{figure*}

\subsection{Runtime Performance}

Since our network architecture is composed of 3D convolutions, we can generate an output prediction in a single forward pass for an input scan, with runtime performance dependent on the 3D volume of the test scene as $\mathcal{O}(\textrm{dimx}\times\textrm{dimy}\times\textrm{dimz})$.
A small scene of size $1.5\times 3.0\times 2.6$ meters ($72\times 152\times 128$ voxels), inference time is $0.33$ seconds; a medium scene of size  $2.8\times 3.9\times 2.6$ meters ($140\times 196\times 128$ voxels) takes $0.86$ seconds, and a large scene of size $6.0\times 6.6\times 2.6$ meters ($300\times 328\times 128$ voxels) takes $2.4$ seconds.

\paragraph{Memory footprint}
During training, our approach operates with a memory footprint of 5.5GB with a batch size of $2$.
At test time, the memory footprint of the small, medium, and large scenes previously mentioned is 0.7GB, 1.7GB, and 6GB respectively.
Very large test scenes can be realized by running our method by chunks of the receptive field size; for instance, our largest test scene spans $34.5\times 49.2\times 2.6$ meters ($1727\times 2461\times 128$ voxels), with a memory footprint of 1.4GB in this fashion.

\subsection{Qualitative Results}

We provide additional qualitative results of colored reconstruction of Matterport3D~\cite{Matterport3D} scans and ShapeNet~\cite{shapenet2015} chairs in Figures~\ref{fig:cmp_mp_more} and \ref{fig:cmp_shapenet_more}, respectively.
As can be seen, our method consistently generates sharper results compared to the baseline methods.
In Figure~\ref{fig:cmp_mp_more}, the comparison to \cite{oechsle2019texture} is shown. Since the approach does not complete geometry, we provide our predicted geometry as input.
In contrast to our method, it is not properly estimating color tones like for the green chair in the bottom row of the figure.
Figure~\ref{fig:cmp_shapenet_more} shows more examples for our experiments on the ShapeNet dataset in comparison to Im2Avatar~\cite{sun2018im2avatar}, PIFu~\cite{saito2019pifu} and Texture Fields~\cite{oechsle2019texture}.

\begin{figure*}[tp]
	\centering
	\includegraphics[width=0.95\linewidth]{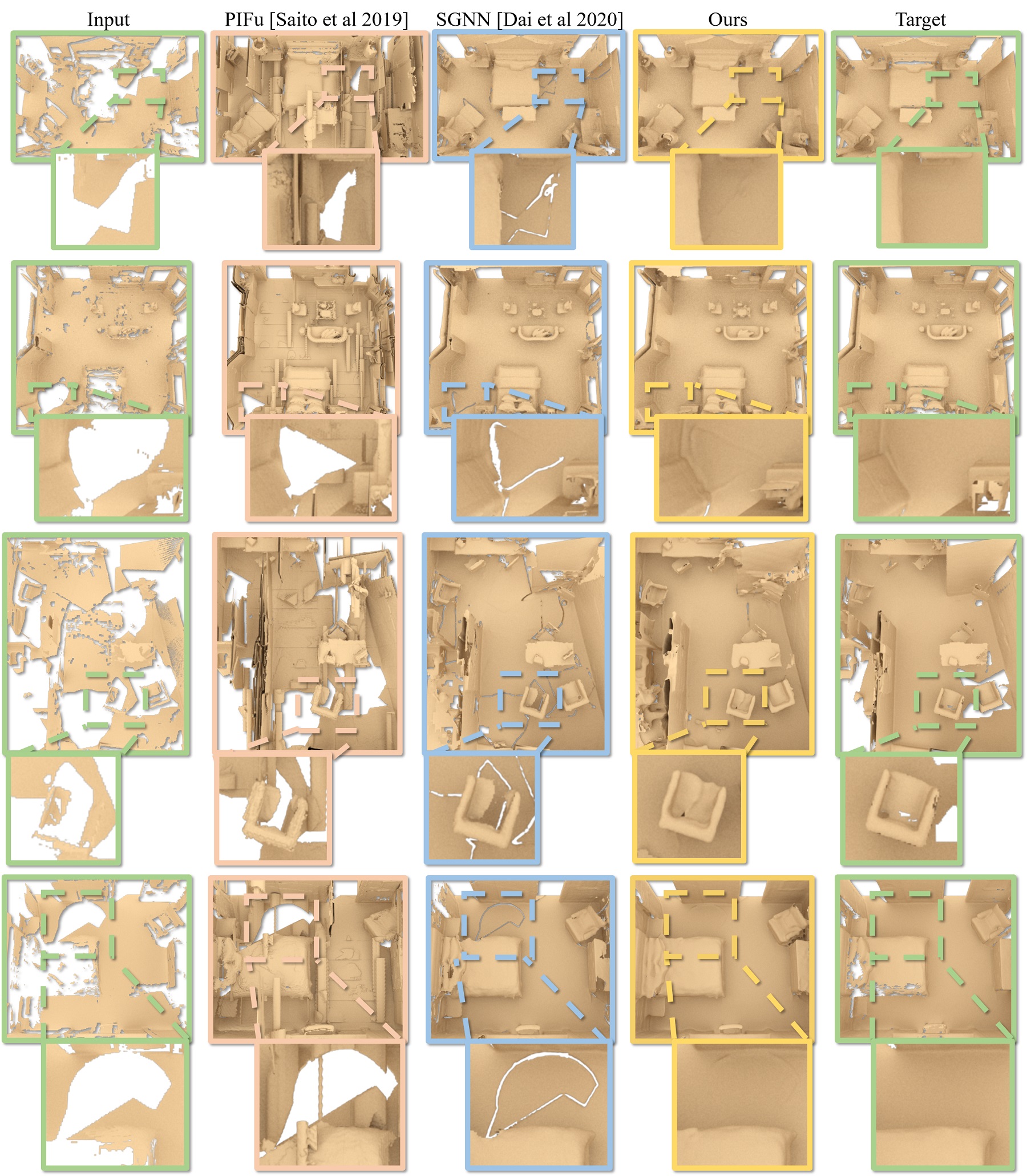}
	\caption{
	Qualitative comparison of geometric completion results on Matterport3D~\cite{Matterport3D} scans. Our view-guided approach mitigates learning from artifacts in the fused 3D reconstruction of the scenes (e.g., small frame misalignments, which can cause seams such as in the SG-NN reconstructions), producing more accurate scene geometry.
    }
	\label{fig:cmp_geo}
\end{figure*}

Additionally, we show qualitative comparisons of the geometric completion of our approach on Matterport3D~\cite{Matterport3D} scans in Figure~\ref{fig:cmp_geo}, in comparison to SG-NN~\cite{dai2020sgnn}.
Both methods were trained on our Matterport3D chunks data, where inputs were composed of $30\%$ of frames available, and targets of $50\%$ of frames available; test target scenes are visualized with all available frames.
The direct 3D supervision guiding SG-NN contains fused errors from small camera estimation misalignments and depth noise (small shifts in the target TSDF), resulting in a tendency to produce a few visible seams in the resulting reconstructions.
In contrast, our view-guided synthesis helps to avoid these artifacts, producing more compelling scene geometry.

%% file: main.bbl
\begin{thebibliography}{10}\itemsep=-1pt

\bibitem{brunet2011mathematical}
Dominique Brunet, Edward~R Vrscay, and Zhou Wang.
\newblock On the mathematical properties of the structural similarity index.
\newblock {\em IEEE Transactions on Image Processing}, 21(4):1488--1499, 2011.

\bibitem{Matterport3D}
Angel~X. Chang, Angela Dai, Thomas~A. Funkhouser, Maciej Halber, Matthias
  Nie{\ss}ner, Manolis Savva, Shuran Song, Andy Zeng, and Yinda Zhang.
\newblock Matterport3d: Learning from {RGB-D} data in indoor environments.
\newblock In {\em 2017 International Conference on 3D Vision, 3DV 2017,
  Qingdao, China, October 10-12, 2017}, pages 667--676, 2017.

\bibitem{shapenet2015}
Angel~X Chang, Thomas Funkhouser, Leonidas Guibas, Pat Hanrahan, Qixing Huang,
  Zimo Li, Silvio Savarese, Manolis Savva, Shuran Song, Hao Su, et~al.
\newblock Shapenet: An information-rich 3d model repository.
\newblock {\em arXiv preprint arXiv:1512.03012}, 2015.

\bibitem{curless1996volumetric}
Brian Curless and Marc Levoy.
\newblock A volumetric method for building complex models from range images.
\newblock In {\em Proceedings of the 23rd annual conference on Computer
  graphics and interactive techniques}, pages 303--312. ACM, 1996.

\bibitem{dai2020sgnn}
Angela Dai, Christian Diller, and Matthias Nie{\ss}ner.
\newblock Sg-nn: Sparse generative neural networks for self-supervised scene
  completion of rgb-d scans.
\newblock In {\em Proc. Computer Vision and Pattern Recognition (CVPR), IEEE},
  2020.

\bibitem{dai2017bundlefusion}
Angela Dai, Matthias Nie{\ss}ner, Michael Zollh{\"{o}}fer, Shahram Izadi, and
  Christian Theobalt.
\newblock Bundlefusion: Real-time globally consistent 3d reconstruction using
  on-the-fly surface reintegration.
\newblock {\em {ACM} Trans. Graph.}, 36(3):24:1--24:18, 2017.

\bibitem{dai2017complete}
Angela Dai, Charles~Ruizhongtai Qi, and Matthias Nie{\ss}ner.
\newblock Shape completion using 3d-encoder-predictor cnns and shape synthesis.
\newblock In {\em 2017 {IEEE} Conference on Computer Vision and Pattern
  Recognition, {CVPR} 2017, Honolulu, HI, USA, July 21-26, 2017}, pages
  6545--6554, 2017.

\bibitem{dai2018scancomplete}
Angela Dai, Daniel Ritchie, Martin Bokeloh, Scott Reed, J{\"{u}}rgen Sturm, and
  Matthias Nie{\ss}ner.
\newblock Scancomplete: Large-scale scene completion and semantic segmentation
  for 3d scans.
\newblock In {\em 2018 {IEEE} Conference on Computer Vision and Pattern
  Recognition, {CVPR} 2018, Salt Lake City, UT, USA, June 18-22, 2018}, pages
  4578--4587.

\bibitem{gatys2016image}
Leon~A Gatys, Alexander~S Ecker, and Matthias Bethge.
\newblock Image style transfer using convolutional neural networks.
\newblock In {\em Proceedings of the IEEE conference on computer vision and
  pattern recognition}, pages 2414--2423, 2016.

\bibitem{goodfellow2014generative}
Ian Goodfellow, Jean Pouget-Abadie, Mehdi Mirza, Bing Xu, David Warde-Farley,
  Sherjil Ozair, Aaron Courville, and Yoshua Bengio.
\newblock Generative adversarial nets.
\newblock In {\em Advances in Neural Information Processing Systems}, pages
  2672--2680, 2014.

\bibitem{heusel2017gans}
Martin Heusel, Hubert Ramsauer, Thomas Unterthiner, Bernhard Nessler, and Sepp
  Hochreiter.
\newblock Gans trained by a two time-scale update rule converge to a local nash
  equilibrium.
\newblock In {\em Advances in neural information processing systems}, pages
  6626--6637, 2017.

\bibitem{huang20173dlite}
Jingwei Huang, Angela Dai, Leonidas~J Guibas, and Matthias Nie{\ss}ner.
\newblock 3dlite: towards commodity 3d scanning for content creation.
\newblock {\em ACM Trans. Graph.}, 36(6):203--1, 2017.

\bibitem{huang2020adversarial}
Jingwei Huang, Justus Thies, Angela Dai, Abhijit Kundu, Chiyu~Max Jiang,
  Leonidas Guibas, Matthias Nie{\ss}ner, and Thomas Funkhouser.
\newblock Adversarial texture optimization from rgb-d scans.
\newblock 2020.

\bibitem{isola2017image}
Phillip Isola, Jun-Yan Zhu, Tinghui Zhou, and Alexei~A Efros.
\newblock Image-to-image translation with conditional adversarial networks.
\newblock In {\em Proceedings of the IEEE conference on computer vision and
  pattern recognition}, pages 1125--1134, 2017.

\bibitem{izadi2011kinectfusion}
Shahram Izadi, David Kim, Otmar Hilliges, David Molyneaux, Richard~A. Newcombe,
  Pushmeet Kohli, Jamie Shotton, Steve Hodges, Dustin Freeman, Andrew~J.
  Davison, and Andrew~W. Fitzgibbon.
\newblock Kinectfusion: real-time 3d reconstruction and interaction using a
  moving depth camera.
\newblock In {\em Proceedings of the 24th Annual {ACM} Symposium on User
  Interface Software and Technology, Santa Barbara, CA, USA, October 16-19,
  2011}, pages 559--568, 2011.

\bibitem{karras2017progressive}
Tero Karras, Timo Aila, Samuli Laine, and Jaakko Lehtinen.
\newblock Progressive growing of gans for improved quality, stability, and
  variation.
\newblock {\em arXiv preprint arXiv:1710.10196}, 2017.

\bibitem{lorensen1987marching}
William~E. Lorensen and Harvey~E. Cline.
\newblock Marching cubes: {A} high resolution 3d surface construction
  algorithm.
\newblock In {\em Proceedings of the 14th Annual Conference on Computer
  Graphics and Interactive Techniques, {SIGGRAPH} 1987, Anaheim, California,
  USA, July 27-31, 1987}, pages 163--169, 1987.

\bibitem{maier2017intrinsic3d}
R. Maier, K. Kim, D. Cremers, J. Kautz, and M. Nie{\ss}ner.
\newblock Intrinsic3d: High-quality {3D} reconstruction by joint appearance and
  geometry optimization with spatially-varying lighting.
\newblock In {\em International Conference on Computer Vision (ICCV)}, Venice,
  Italy, October 2017.

\bibitem{maturana2015voxnet}
Daniel Maturana and Sebastian Scherer.
\newblock Voxnet: {A} 3d convolutional neural network for real-time object
  recognition.
\newblock In {\em 2015 {IEEE/RSJ} International Conference on Intelligent
  Robots and Systems, {IROS} 2015, Hamburg, Germany, September 28 - October 2,
  2015}, pages 922--928, 2015.

\bibitem{OccupancyNetworks}
Lars Mescheder, Michael Oechsle, Michael Niemeyer, Sebastian Nowozin, and
  Andreas Geiger.
\newblock Occupancy networks: Learning 3d reconstruction in function space.
\newblock In {\em Proceedings IEEE Conf. on Computer Vision and Pattern
  Recognition (CVPR)}, 2019.

\bibitem{newcombe2011kinectfusion}
Richard~A. Newcombe, Shahram Izadi, Otmar Hilliges, David Molyneaux, David Kim,
  Andrew~J. Davison, Pushmeet Kohli, Jamie Shotton, Steve Hodges, and Andrew~W.
  Fitzgibbon.
\newblock Kinectfusion: Real-time dense surface mapping and tracking.
\newblock In {\em 10th {IEEE} International Symposium on Mixed and Augmented
  Reality, {ISMAR} 2011, Basel, Switzerland, October 26-29, 2011}, pages
  127--136, 2011.

\bibitem{oechsle2019texture}
Michael Oechsle, Lars Mescheder, Michael Niemeyer, Thilo Strauss, and Andreas
  Geiger.
\newblock Texture fields: Learning texture representations in function space.
\newblock In {\em Proceedings of the IEEE International Conference on Computer
  Vision}, pages 4531--4540, 2019.

\bibitem{Park2019DeepSDFLC}
Jeong~Joon Park, Peter Florence, Julian Straub, Richard~A. Newcombe, and Steven
  Lovegrove.
\newblock Deepsdf: Learning continuous signed distance functions for shape
  representation.
\newblock In {\em {IEEE} Conference on Computer Vision and Pattern Recognition,
  {CVPR} 2019, Long Beach, CA, USA, June 16-20, 2019}, pages 165--174, 2019.

\bibitem{peng2020convolutional}
Songyou Peng, Michael Niemeyer, Lars Mescheder, Marc Pollefeys, and Andreas
  Geiger.
\newblock Convolutional occupancy networks.
\newblock {\em arXiv preprint arXiv:2003.04618}, 2, 2020.

\bibitem{riegler2017OctNet}
Gernot Riegler, Ali~Osman Ulusoy, and Andreas Geiger.
\newblock Octnet: Learning deep 3d representations at high resolutions.
\newblock In {\em 2017 {IEEE} Conference on Computer Vision and Pattern
  Recognition, {CVPR} 2017, Honolulu, HI, USA, July 21-26, 2017}, pages
  6620--6629, 2017.

\bibitem{saito2019pifu}
Shunsuke Saito, Zeng Huang, Ryota Natsume, Shigeo Morishima, Angjoo Kanazawa,
  and Hao Li.
\newblock Pifu: Pixel-aligned implicit function for high-resolution clothed
  human digitization.
\newblock In {\em Proceedings of the IEEE International Conference on Computer
  Vision}, pages 2304--2314, 2019.

\bibitem{simonyan2014very}
Karen Simonyan and Andrew Zisserman.
\newblock Very deep convolutional networks for large-scale image recognition.
\newblock {\em arXiv preprint arXiv:1409.1556}, 2014.

\bibitem{song2017ssc}
Shuran Song, Fisher Yu, Andy Zeng, Angel~X. Chang, Manolis Savva, and Thomas~A.
  Funkhouser.
\newblock Semantic scene completion from a single depth image.
\newblock In {\em 2017 {IEEE} Conference on Computer Vision and Pattern
  Recognition, {CVPR} 2017, Honolulu, HI, USA, July 21-26, 2017}, pages
  190--198, 2017.

\bibitem{sun2018im2avatar}
Yongbin Sun, Ziwei Liu, Yue Wang, and Sanjay~E Sarma.
\newblock Im2avatar: Colorful 3d reconstruction from a single image.
\newblock {\em arXiv preprint arXiv:1804.06375}, 2018.

\bibitem{szegedy2016rethinking}
Christian Szegedy, Vincent Vanhoucke, Sergey Ioffe, Jon Shlens, and Zbigniew
  Wojna.
\newblock Rethinking the inception architecture for computer vision.
\newblock In {\em Proceedings of the IEEE conference on computer vision and
  pattern recognition}, pages 2818--2826, 2016.

\bibitem{whelan2015elasticfusion}
Thomas Whelan, Stefan Leutenegger, Renato~F. Salas{-}Moreno, Ben Glocker, and
  Andrew~J. Davison.
\newblock Elasticfusion: Dense {SLAM} without {A} pose graph.
\newblock In {\em Robotics: Science and Systems XI, Sapienza University of
  Rome, Rome, Italy, July 13-17, 2015}, 2015.

\bibitem{wu20153d}
Zhirong Wu, Shuran Song, Aditya Khosla, Fisher Yu, Linguang Zhang, Xiaoou Tang,
  and Jianxiong Xiao.
\newblock 3d shapenets: {A} deep representation for volumetric shapes.
\newblock In {\em {IEEE} Conference on Computer Vision and Pattern Recognition,
  {CVPR} 2015, Boston, MA, USA, June 7-12, 2015}, pages 1912--1920, 2015.

\bibitem{NIPS2019_8340}
Qiangeng Xu, Weiyue Wang, Duygu Ceylan, Radomir Mech, and Ulrich Neumann.
\newblock Disn: Deep implicit surface network for high-quality single-view 3d
  reconstruction.
\newblock In H. Wallach, H. Larochelle, A. Beygelzimer, F. Alch\'{e}-Buc, E.
  Fox, and R. Garnett, editors, {\em Advances in Neural Information Processing
  Systems 32}, pages 492--502. 2019.

\bibitem{yang2019pointflow}
Guandao Yang, Xun Huang, Zekun Hao, Ming-Yu Liu, Serge Belongie, and Bharath
  Hariharan.
\newblock Pointflow: 3d point cloud generation with continuous normalizing
  flows.
\newblock In {\em The IEEE International Conference on Computer Vision (ICCV)},
  October 2019.

\bibitem{reco_star}
Michael Zollhöfer, Patrick Stotko, Andreas Görlitz, Christian Theobalt,
  Matthias Nießner, Reinhard Klein, and Andreas Kolb.
\newblock State of the art on 3d reconstruction with rgb‐d cameras.
\newblock {\em Computer Graphics Forum}, 37:625--652, 05 2018.

\bibitem{zollhoefer2015shading}
Michael Zollh{\"o}fer, Angela Dai, Matthias Innmann, Chenglei Wu, Marc
  Stamminger, Christian Theobalt, and Matthias Nie{\ss}ner.
\newblock Shading-based refinement on volumetric signed distance functions.
\newblock {\em ACM Transactions on Graphics (TOG)}, 2015.

\end{thebibliography}
